\documentclass[11pt]{article}

\usepackage[utf8]{inputenc}
\usepackage[T1]{fontenc}
\usepackage{lmodern}
\usepackage{textcomp}
\usepackage{microtype}

\usepackage[margin=1in]{geometry}
\usepackage{setspace}
\usepackage{graphicx}
\usepackage{booktabs}
\usepackage{array}
\usepackage{enumitem}

\usepackage{url}
\usepackage[hidelinks, breaklinks]{hyperref}
\usepackage[round,authoryear]{natbib}

\usepackage{titling}
\pretitle{\begin{center}\LARGE\bfseries}
\posttitle{\par\end{center}\vskip 0.4em}

\preauthor{\begin{center}\normalsize}
\postauthor{\end{center}\vskip 0.2em}

\predate{}
\postdate{}

\title{Structural Silence: When AI Infrastructure Fails Speakers of Underrepresented Languages\thanks{Presented at the 69th Annual Conference of the International Linguistic Association (ILA), New York, NY, April 30--May 2, 2026.} \\
\large A Case Study in Bengali, the Digital Divide, and the Hidden Cost of English-Centric Design}

\author{
\begin{tabular}{c}
Avijit Roy\\
\small John Jay College of Criminal Justice, CUNY\\
\small \texttt{aroy@jjay.cuny.edu}\\[0.5em]
Proma Roy\\
\small The City College of New York, CUNY\\
\small \texttt{proma.roy33@stu-mail.ccny.cuny.edu}
\end{tabular}
}

\date{}

\begin{document}

\begin{singlespace}
\maketitle
\end{singlespace}

\begin{abstract}
\small
\setstretch{1.1}
Artificial intelligence tools for education and language support are increasingly framed as scalable responses to access gaps in under-resourced communities. Yet the infrastructure underlying these tools---training corpora, tokenization schemes, evaluation benchmarks, and deployment architectures---encodes a set of assumptions that systematically disadvantages speakers of underrepresented languages before a single model is trained.

This paper examines those assumptions through the lens of Bengali, one of the world's most widely spoken languages with roughly 285 million speakers \citep{ethnologue2025, icls2026}, and the structural barriers that emerge when attempting to build AI-assisted educational tools for Bengali-speaking learners in low-connectivity environments. We identify four interlocking structural failures: a severe web presence gap (Bengali accounts for fewer than 0.5\% of global web content despite representing nearly 4\% of the global population) \citep{pimienta2024, w3techs2026}; a 67:1 training token deficit between English and Bengali in major multilingual corpora \citep{khan2024sangraha, langlais2025commoncorpus}; a tokenization penalty rooted in Bengali's alphasyllabary script that compounds the data deficit by requiring higher token fertility rates \citep{shahriar2024improving}; and a connectivity exclusion that renders cloud-dependent AI tools inaccessible to the rural populations most likely to benefit from them, where individual internet penetration stands at 36.5\% compared to 71.4\% in urban areas \citep{bbs2025ictsurvey, dailystar2025internet}.

These failures reflect recurring patterns in infrastructure design that cannot be resolved through isolated technical adjustments alone. They are the downstream consequences of longstanding resource allocation decisions, institutional priorities, and design defaults that did not center certain languages in mainstream AI development. Dataset scarcity should therefore be understood as a structural barrier shaped by those decisions rather than as an individual researcher limitation. Offline-first design functions as an equity-oriented infrastructure strategy rather than a secondary technical compromise. We close with specific directions for how the linguistics and AI communities might respond.

\vspace{0.3em}
\end{abstract}

\section{Introduction}

Programming education is positioned, with increasing frequency, as a pathway to economic participation in technology-driven societies. Alongside it, AI-assisted educational tools are positioned as scalable solutions to the uneven distribution of qualified instruction---tools that can, in principle, provide debugging assistance, concept explanation, and interactive feedback to learners who have no access to expert teachers.

That premise has considerable appeal, but the infrastructure underlying these tools does not support it uniformly across languages. When a developer or researcher attempts to build an AI-assisted educational tool for Bengali-speaking learners, they encounter a sequence of interconnected failures rooted in how AI infrastructure was built and who it was built for. These failures are not incidental byproducts of early-stage technology. They are structural features of the training corpora that models learn from, the tokenizers that process Bengali text, the benchmarks used to evaluate model performance, and the deployment architectures that assume reliable internet connectivity. Each failure also intensifies the effects of the others over time. Together, they constitute what we call \emph{structural silence}---the systematic exclusion of a language from AI infrastructure not through explicit policy but through the accumulated weight of design decisions that were never made with that language in mind.

Bengali is a revealing case study precisely because its scale forecloses the most convenient explanations. It is among the most widely spoken languages in the world \citep{ethnologue2025, icls2026}, with a literary history spanning more than 1,400 years. It is the official language of Bangladesh and one of the scheduled languages of India. UNESCO recognizes International Mother Language Day on 21 February, an observance rooted in the historical struggle over linguistic recognition in Bengal \citep{unesco_imld, un_imld}. In 2024, the Government of India conferred Classical Language status on Bengali \citep{pmindia2024classical}. Bengali's marginal position in AI infrastructure therefore cannot be attributed to demographic insignificance or a limited written tradition. It reflects cumulative patterns of resource allocation, institutional priority-setting, and design assumptions that centered other languages---primarily English---in the development of contemporary AI systems.

This paper is framed as a case study and analytic synthesis rather than as a new benchmark or systems contribution. Its purpose is to consolidate technical, infrastructural, and educational evidence into a single explanatory account of why AI support remains uneven for Bengali-speaking learners, and to identify the structural logic by which that unevenness is reproduced.

The paper proceeds as follows. Section~\ref{sec:background} reviews the relevant background on low-resource language NLP and AI infrastructure. Section~\ref{sec:failures} identifies and evidences four interlocking structural failures. Section~\ref{sec:doubleburden} connects those failures to the cognitive and educational consequences for Bengali-speaking learners. Section~\ref{sec:discussion} discusses implications for the linguistics and AI research communities. Section~\ref{sec:conclusion} concludes.

\section{Background}
\label{sec:background}

\subsection{The Language Diversity Problem in NLP}

The distribution of resources across languages in natural language processing has been uneven since the field's inception, but the scale of that inequality became newly visible with the emergence of large language models. \citet{joshi2020state} provided a landmark taxonomy of this inequality, classifying languages into six categories from ``winners'' (those with abundant data and research attention) to ``left-behinds'' (those with no meaningful digital presence). Their analysis found that only a small share of the world's languages could be classified as winners in NLP resource terms, while the overwhelming majority were left behind \citep{joshi2020state}. Bengali occupies an intermediate position---it has received some research attention, particularly in tasks like sentiment analysis, machine translation, and named entity recognition---but remains substantially underserved in technical domains, particularly code understanding and educational NLP.

The reasons for this distribution are well understood. Training data for large language models is largely harvested from the web, so languages with greater online presence receive more training data. Greater data volumes generally produce better-performing models, which attract more users, more research investment, and further development. The cycle is self-reinforcing, and it began from a starting position heavily weighted toward English.

\subsection{Bengali in NLP Research}

Bengali NLP has grown considerably as a research area over the past decade. Resources including BanglaBERT \citep{bhattacharjee2022banglabert}, XL-Sum \citep{hasan2021xlsum}, and the Sangraha multilingual corpus \citep{khan2024sangraha} represent meaningful investments in Bengali language infrastructure. The BLP shared tasks at ACL/AACL have also helped standardize evaluation for Bengali NLP, including the 2025 task specifically targeting code generation in Bangla \citep{raihan2025blp}.

Despite these advances, the BenLLM-Eval benchmark \citep{kabir2024benllm} and subsequent evaluations \citep{bhowmik2025evaluating} consistently show that general-purpose large language models---including those marketed as multilingual---perform substantially worse on Bengali downstream tasks than on equivalent English tasks. This performance gap varies across models and tasks, but it appears consistently enough to treat it as a feature of the current infrastructure rather than as an artifact of any single evaluation design.

\subsection{Cognitive Load and Language of Instruction}

Cognitive Load Theory \citep{sweller2011clt} provides a well-established theoretical framework for understanding why language of instruction matters in technical education. The theory holds that human working memory has limited capacity, and that learning outcomes depend on how effectively instructional design manages the total cognitive load placed on that capacity. When learners must process new technical concepts through a second language, the linguistic processing demands compete directly with the conceptual processing demands---with predictable consequences for retention and comprehension.

\citet{roussel2017cognitiveload} demonstrated this empirically in higher education contexts, finding that exposure to academic content in a foreign language without explicit language support significantly reduced both content and language learning outcomes compared to native-language instruction, even when the material was otherwise identical. Studies specifically examining programming education for non-native English speakers \citep{soosairaj2018native, dodoo2025teacher} find consistent evidence that linguistic barriers compound the already substantial cognitive demands of learning to program, particularly in error diagnosis and debugging tasks.

\subsection{Offline-First Computing and Educational Equity}

The framing of internet connectivity as a prerequisite for AI tool access has received increasing scrutiny in computing and sustainable societies research. The ACM COMPASS community has documented persistent rural--urban digital divides in Bangladesh and comparable contexts \citep{rashed2025bridging}, noting that household-level connectivity statistics often obscure the reality of individual access: a household with one shared smartphone may count as connected while the individual learner within it has no reliable access to bandwidth-intensive applications.

Quantization techniques and parameter-efficient fine-tuning methods \citep{dettmers2023qlora, hu2022lora} have made local inference on consumer-grade hardware increasingly feasible, opening a path to offline-first educational AI deployment that was not available even five years ago.

\section{Four Structural Failures}
\label{sec:failures}

\subsection{Failure One: The Web Presence Gap}

The first structural failure precedes AI development itself: it concerns the distribution of language on the web, which remains the primary source of large-scale training data.

AI training corpora are predominantly assembled from web crawls. Common Crawl, the most widely used source for large model pre-training, indexes content in rough proportion to its online presence. A language's representation in training data is therefore constrained by its representation on the web, and Bengali's web presence is severely disproportionate to its speaker population.

Bengali speakers represent roughly 4\% of the global population---approximately 242 million native speakers out of a world population of about 8 billion \citep{ethnologue2025, icls2026}---yet Bengali accounts for fewer than 0.5\% of global web content \citep{pimienta2024, icls2026}. English, by contrast, accounts for approximately 49.5\% of global web content \citep{w3techs2026}, despite English native speakers representing a share of the global population broadly comparable in scale to Bengali's.

\begin{figure}[t]
  \centering
  \includegraphics[width=\linewidth]{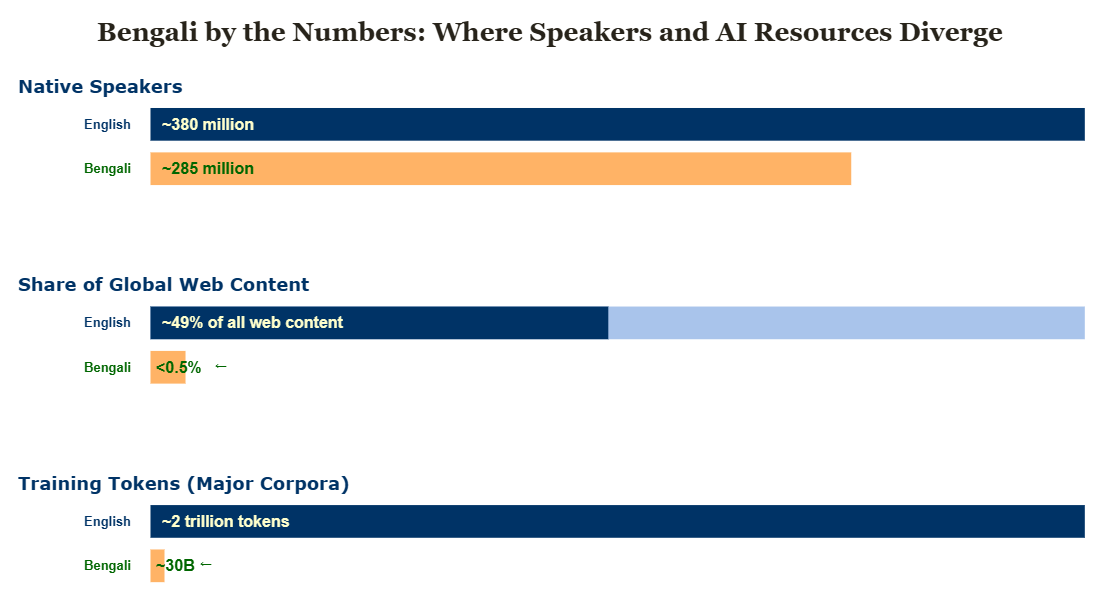}
  \caption{Speaker-resource divergence for Bengali and English.}
  \label{fig:speaker_resource}
\end{figure}

This disparity is not primarily a function of internet access rates. It reflects the economics of content production, the historical concentration of technology development in English-speaking contexts, and the absence of sustained institutional investment in Bengali digital content creation. The result is that before any model is trained and before any dataset is curated, Bengali enters the AI development pipeline from a position of severe disadvantage in the raw material that large-scale language modeling requires.

\subsection{Failure Two: The Training Token Deficit}

The web presence gap translates directly into a training token deficit that is quantifiable and substantial.

The Sangraha corpus, one of the largest multilingual Indic language resources documented in the ACL literature, contains roughly 251 billion tokens across 22 Indic languages \citep{khan2024sangraha}. The project allocates approximately 30 billion tokens to Bengali, while large English corpora such as Common Corpus reach approximately 2 trillion English tokens \citep{khan2024sangraha, langlais2025commoncorpus}. The ratio between English and Bengali token availability in major training resources is therefore approximately \textbf{67:1}.\footnote{This ratio is derived by comparing the Bengali token allocation in Sangraha (~30B tokens; \citealt{khan2024sangraha}) against the approximate English token count in Common Corpus (~2T tokens; \citealt{langlais2025commoncorpus}). These are not the same corpus, and the comparison is intended to illustrate the order-of-magnitude disparity in available training data rather than an exact within-corpus measurement.}

This ratio carries direct consequences for model performance. \citet{kabir2024benllm} evaluated multiple large language models on Bengali NLP tasks and found consistent performance gaps relative to English equivalents. A 2025 evaluation of ten open-source LLMs across Bengali and English benchmarks likewise showed that Bengali performance was lower across model families, particularly for smaller models like Mistral, and that the gaps correlated with tokenization efficiency and model scale in ways consistent with data-volume explanations \citep{bhowmik2025evaluating}.

Inclusion in a multilingual training set does not, on its own, resolve this deficit. \citet{chang2024goldfish} showed that large multilingual models such as XGLM (4.5B parameters) and BLOOM (7.1B parameters) can underperform even simple bigram baselines on low-resource language perplexity benchmarks. Multilingual training coverage does not guarantee multilingual competence, and the gap between them tends to widen for languages with small shares of training data.

\begin{figure}[t]
  \centering
  \includegraphics[width=0.3\linewidth]{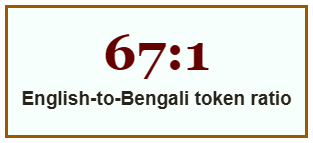}
  \caption{The 67:1 English--Bengali training token gap.}
  \label{fig:token_gap}
\end{figure}

\subsection{Failure Three: The Tokenization Penalty}

The third failure operates at the level of the tokenization pipeline and is less immediately visible than the data volume disparity, though its consequences compound the data deficit in important ways. Because of how standard tokenizers process Bengali script, equal data volumes do not translate to equal representational quality.

Standard large language model tokenizers---including Byte Pair Encoding (BPE) and WordPiece---were designed and optimized for Latin-script languages. Bengali employs an alphasyllabary script in which base characters are frequently modified by diacritics (\emph{matras}) and conjunct forms (\emph{yuktakshar}) that alter pronunciation and meaning. These modifications can occur on either side of a base character, forming multi-character grapheme clusters that do not decompose cleanly under standard tokenization schemes \citep{shahriar2024improving}.

\begin{figure}[t]
  \centering
  \includegraphics[width=\linewidth]{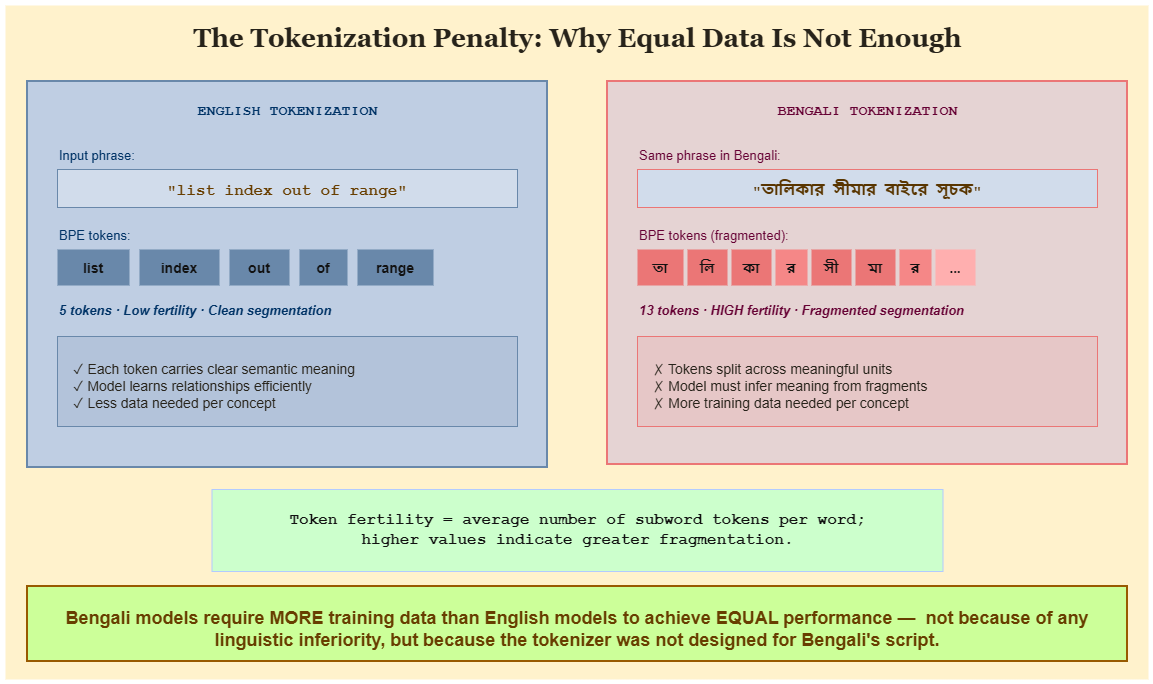}
  \caption{Illustrative comparison of tokenization fertility between English and Bengali scripts.}
  \label{fig:tokenization_penalty}
\end{figure}

The practical consequence is elevated token fertility: Bengali text requires significantly more subword tokens to represent the same semantic content as comparable English text. Token fertility refers to the average number of subword tokens required to represent a single word or linguistic unit; higher values indicate greater fragmentation during tokenization. This fragmentation increases computational overhead and, more substantially, disrupts the linguistic units that models rely on to build representations. A model trained on heavily fragmented Bengali input must learn inter-token relationships from structurally noisier data than an English model working from comparatively clean, low-fertility tokenization.\footnote{Token fertility values vary by tokenizer and corpus. \citet{shahriar2024improving} document substantially higher fertility rates for Bengali than English under standard BPE tokenization. Figure~\ref{fig:tokenization_penalty} presents an illustrative example; exact fertility ratios depend on tokenizer vocabulary size and training data composition.}

The combined effect is cumulative and compounding. The training token deficit already places Bengali at a disadvantage in terms of data volume. The tokenization penalty means that the tokens Bengali does receive are processed less efficiently than English tokens---so parity in data volume would still leave a performance gap. Equitable Bengali model performance therefore requires not just proportional data, but substantially more data to compensate for structural inefficiencies in how that data is handled by tokenizers built for a different script.

\subsection{Failure Four: The Connectivity Exclusion}

The fourth failure is architectural, concerning where AI tools are deployed and under what connectivity assumptions they operate.

The dominant paradigm for AI tool deployment is cloud-based: a user submits a query to a remote server, the server runs inference, and the result is returned over a network connection. For learners in rural Bangladesh, the conditions this architecture requires are frequently absent.

According to the Bangladesh Bureau of Statistics ICT Access and Use Survey (FY2024--25), individual internet penetration in rural Bangladesh stands at 36.5\%, compared to 71.4\% in urban areas \citep{bbs2025ictsurvey, dailystar2025internet}. Contemporary reporting on the same survey documents that national household internet access sits just above 50\%, with only 47.2\% of the population counted as internet users by the end of 2024 \citep{bbs2025ictsurvey, tbs2025internet}. For the learners under consideration here, these figures constitute a material constraint on access to any tool that requires sustained network connectivity.

The household-level statistics, though slightly less stark, obscure a further distinction: household access is not individual access. The Business Standard reports that only 9.2\% of Bangladeshi households own a computer, with just 9\% of individuals personally using one \citep{tbs2025internet}. In a household with one shared smartphone, every member may be counted in household connectivity statistics, but only those with physical access to the device---and a sufficient data budget---can use it for extended learning. The cost dimension is relevant here as well: \emph{The Daily Star} reports that supplementary duty on mobile data in Bangladesh increased from 3\% in FY2016 to 23\% in 2024, meaning that rural students in lower-income households now face a substantially higher real cost for the data that cloud-dependent tools consume \citep{dailystar2025internet}.

For these learners, an AI tutoring system that depends on a stable internet connection is not merely less convenient than a local alternative. It is functionally inaccessible, and its design implicitly directs its benefits toward the connected, urban populations who are already better served by existing educational infrastructure.

\begin{figure}[t]
  \centering
  \includegraphics[width=0.95\linewidth]{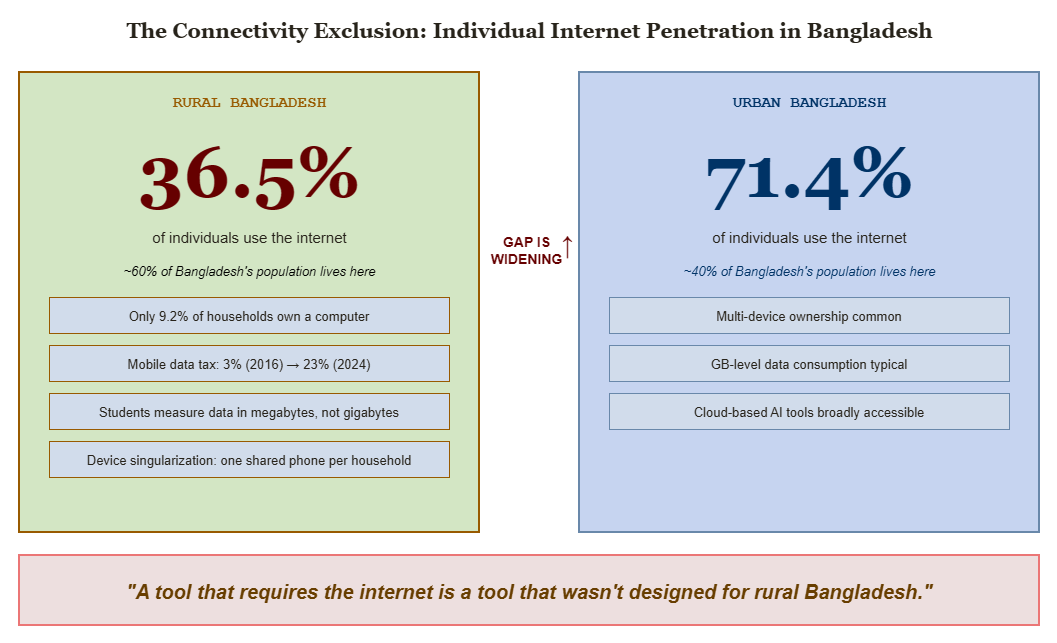}
  \caption{Connectivity exclusion in rural versus urban Bangladesh.}
  \label{fig:connectivity_exclusion}
\end{figure}

\section{Educational Consequences: The Double Burden}
\label{sec:doubleburden}

The four structural failures described above converge in a concrete educational outcome: Bengali-speaking learners who attempt to use AI-assisted programming tools encounter a compounded set of cognitive demands that their English-speaking counterparts do not face.

Learning to program is cognitively demanding regardless of the learner's first language. Debugging in particular---identifying why code fails, tracing the root cause, and formulating a correction---places substantial demands on working memory, requiring the simultaneous maintenance of multiple representations: the intended behavior, the actual behavior, the error message, and the corrective action.

When the AI tool's explanations are delivered in English---as they will be for any general-purpose model, given the training data distribution described above---a learner with limited English proficiency must process both the linguistic content of the explanation and its technical content concurrently. According to Cognitive Load Theory \citep{sweller2011clt}, the combined load of language processing and conceptual processing frequently exceeds working memory capacity, reducing both retention and the learner's ability to apply new knowledge to novel problems.

\citet{roussel2017cognitiveload} demonstrated this experimentally with 294 higher education students, showing that content presented in a foreign language produced significantly lower outcomes across both content and language learning compared to native-language presentation, under conditions where all other instructional variables were held constant. \citet{soosairaj2018native} found analogous patterns in programming education, with bilingual instruction (English and Tamil) outperforming English-only instruction for non-English-dominant students on linked-list concept learning. These findings suggest that the language of an AI tutor's output is not a minor formatting choice. For learners without strong English proficiency, it determines whether the explanation is cognitively accessible at all.

\begin{figure}[t]
  \centering
  \includegraphics[width=\linewidth]{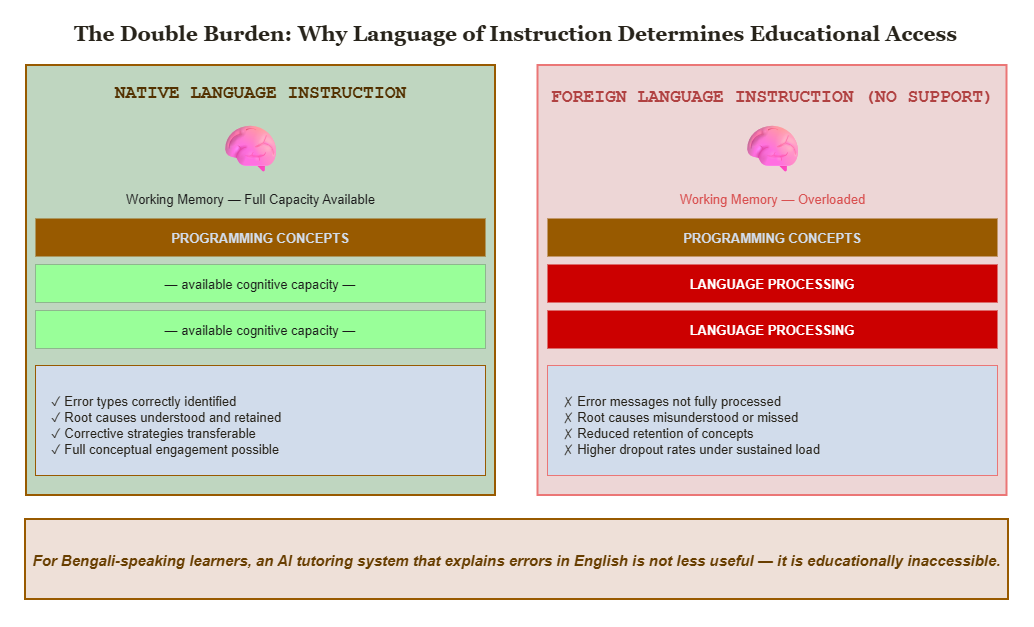}
  \caption{Illustration of the dual cognitive burden faced by learners processing technical content through a second language.}
  \label{fig:double_burden}
\end{figure}

An AI tutoring system that explains Python errors in Bengali therefore serves a function well beyond localization or convenience. For a rural Bengali-speaking student, native-language instruction is a prerequisite for meaningful educational access rather than an enhancement layered on top of it.

\section{Discussion}
\label{sec:discussion}

\subsection{Reframing Dataset Scarcity}

The dominant narrative in low-resource language NLP frames dataset scarcity as a technical problem awaiting a technical solution: there is insufficient data, and the remedy is to produce more. This framing captures part of the situation but does not capture enough of it. Treating scarcity purely as a technical gap can obscure its structural origins, including patterns of funding, publication incentives, and research priorities that have historically directed resources toward high-resource languages.

When a researcher building Bengali AI tools must construct foundational datasets from scratch---curating training examples, validating translations, developing evaluation benchmarks, establishing annotation protocols---they are not compensating for gaps in their own preparation. They are performing infrastructure work that the field has historically declined to fund as a primary contribution or to credit on equal terms with model development. The absence of a Bengali programming education corpus should not be interpreted as a natural scarcity. It reflects the cumulative effect of institutional decisions about what kinds of research outputs deserve recognition and support.

This distinction directly affects how resources are allocated going forward. Researchers who treat low-resource language infrastructure as secondary work---necessary but preliminary, worth doing but not worth publishing on its own terms---reproduce the same asymmetry that manifests in model performance later. Recognizing dataset construction for underrepresented languages as a substantive research contribution, comparable to architectural innovation, is likely to be an important condition for changing that pattern. Publication norms that routinely credit benchmark creation, corpus development, and annotation methodology for high-resource languages should extend the same recognition to equivalent work in low-resource settings.

The analysis in this paper draws on published benchmarks, public infrastructure indicators, and established educational theory rather than a newly collected learner dataset. Its contribution is therefore explanatory and diagnostic: to clarify the infrastructure-level logic by which language inequity in AI support is produced and reproduced, rather than to introduce new empirical data.

\subsection{Offline-First as Equity Architecture}

Offline-first deployment is often described in the literature as a technical accommodation for poor connectivity---a compromise that accepts reduced capability in exchange for broader reach. That framing can mischaracterize the design choice involved and may shift attention away from its broader social implications.

A system designed for cloud deployment operates under a connectivity assumption that excludes a substantial share of the population described in Section~\ref{sec:failures}. Designing for local, offline-capable deployment removes that assumption and extends the system's potential reach to users who would otherwise be excluded. Given the connectivity distribution documented above, this architectural choice represents a different prioritization of which users the system is intended to serve rather than a degraded version of cloud deployment.

There is also a sustainability dimension worth noting. Local inference on a quantized 7-billion-parameter model consumes a fraction of the energy required to route each query through a remote cloud API. At institutional scale---a school computer lab serving dozens of students per day---the cumulative difference in energy consumption and infrastructure cost is material. Lightweight local models thus offer advantages along both equity and sustainability axes, which is consistent with the ACM COMPASS community's emphasis on computing design that accounts for social and environmental consequences simultaneously.

In practical terms, educational AI systems intended for low-connectivity settings should be evaluated not only for response quality under ideal conditions, but also for whether they remain usable under the bandwidth, device, and cost constraints that characterize the communities they are intended to serve. Evaluation frameworks that test only high-connectivity performance implicitly validate tools that are inaccessible to large portions of their stated target populations.

\subsection{The Role of the Linguistics Community}

Linguists have the critical vocabulary to identify and name the assumptions embedded in AI infrastructure that AI developers---who have been trained primarily in high-resource language contexts---often lack the disciplinary background to see. Bengali's tokenization penalty, for example, is a consequence of applying a tokenization paradigm developed for languages with relatively simple grapheme-to-phoneme correspondences and low morphological complexity to a language with a rich, structurally distinct writing system. Recognizing it as such, and articulating precisely why standard BPE is poorly suited to the modification structure of Bengali, is exactly the kind of analysis that linguistic expertise makes possible.

More broadly, the English-centric assumptions embedded in AI development pipelines are, at their root, assumptions about whose linguistic structures are legible to standard tools, whose orthographic conventions are treated as defaults, and whose community needs figure in system design decisions. Sociolinguistics, the critical study of language, and linguistic typology all have sophisticated frameworks for analyzing questions of this kind. Applying those frameworks to AI infrastructure is not peripheral to the discipline's concerns. Given the scale at which AI systems now mediate language use, language learning, and language production, it may be among its most consequential current applications.

For that reason, the linguistics community should not treat AI infrastructure debates as adjacent to its core work. It is one of the few fields positioned to name the places where apparently technical defaults function as language-specific assumptions with unequal consequences across speech communities.

\section{Conclusion}
\label{sec:conclusion}

This paper has identified four recurring barriers that compound to disadvantage Bengali-speaking learners in AI-assisted educational settings: a web presence gap, a training token disparity, a tokenization penalty, and a connectivity exclusion. Taken together, these barriers suggest that uneven model performance across languages is unlikely to be explained by model quality alone. A more informative interpretation is that current performance distributions reflect long-standing patterns in how data resources, evaluation practices, and the infrastructure decisions that shaped deployment assumptions have been distributed across languages---decisions that were not neutral and whose effects have accumulated over time.

The implications extend beyond Bengali, though Bengali serves as a particularly useful case because of its demographic scale, literary history, and growing digital presence. A language with hundreds of millions of speakers, classical recognition, and an active NLP research community still confronts substantial gaps across key layers of AI infrastructure. The structural challenges facing smaller and less-resourced languages are likely to be considerably more severe.

This paper's contribution is therefore not a new benchmark or deployment system, but a clarification of the infrastructure-level logic that helps explain why language inequity persists in multilingual AI. Three responses follow from that analysis. First, low-resource language infrastructure work---datasets, benchmarks, and evaluation protocols---warrants recognition as primary research, not as supporting labor that precedes more substantive contributions. Second, offline-capable design should be treated as a serious educational access strategy with its own design requirements and evaluation criteria, appropriate for contexts where connectivity cannot be assumed. Third, linguistic analysis should play a central role in identifying the tokenization, evaluation, and modeling assumptions that remain invisible when English is treated as the reference standard.

The persistence of language inequity in AI systems reflects less a limitation in modeling techniques than the uneven availability of large-scale training data across languages. Modern AI pipelines depend on massive token volumes, and languages with limited digital presence cannot easily meet those requirements. As a result, performance differences often reflect differences in training feasibility rather than differences in linguistic complexity or user demand. Addressing that inequity will therefore require sustained institutional commitment to expanding dataset development for underrepresented languages and supporting infrastructure that makes large-scale training feasible.

\section*{Acknowledgments}
This work was presented as a poster at the 69th Annual Conference of the International Linguistic Association (ILA), \textit{Beyond Human Language: The Impact of AI on Linguistics}, John Jay College of Criminal Justice, CUNY, April 30 -- May 2, 2026. The authors thank Professor Vivek Sharma for the conference introduction and the ILA reviewers for their valuable feedback.

\section*{Generative AI Disclosure}
The authors used AI-assisted tools for grammar checking and literature search assistance during the preparation of this manuscript. All research framing, arguments, analysis, and conclusions are the original work of the authors.

\bibliographystyle{plainnat}
\bibliography{structural_silence_refs}

@inproceedings{bhattacharjee2022banglabert,
  title     = {{BanglaBERT}: Language Model Pretraining and Benchmarks for Low-Resource Language Understanding Evaluation in {Bangla}},
  author    = {Bhattacharjee, Abhik and Hasan, Tahmid and Islam, Wasi Ahmad and Mubasshir, Kazi Samin and Li, Yuan-Fang and Kang, Yong-Bin and Rahman, M. Sohel and Shahriyar, Rifat},
  booktitle = {Findings of the Association for Computational Linguistics: NAACL 2022},
  pages     = {1863--1881},
  year      = {2022},
  publisher = {Association for Computational Linguistics},
  url       = {https://aclanthology.org/2022.findings-naacl.98/}
}

@article{bhowmik2025evaluating,
  title   = {Evaluating {LLMs}' Multilingual Capabilities for {Bengali}: Benchmark Creation and Performance Analysis},
  author  = {Bhowmik, Shimanto and Dipto, Tawsif Tashwar and Islam, Md Sazzad and Hsu, Sheryl and Reasat, Tahsin},
  journal = {arXiv preprint arXiv:2507.23248},
  year    = {2025},
  url     = {https://arxiv.org/abs/2507.23248}
}

@misc{bbs2025ictsurvey,
  title        = {{ICT Access and Use Survey 2024--25}},
  author       = {{Bangladesh Bureau of Statistics}},
  year         = {2025},
  howpublished = {Government Survey Report},
  publisher    = {Government of the People's Republic of Bangladesh},
  url          = {https://bbs.gov.bd/site/page/b588b454-0f88-4679-bf20-90e06dc1d10b/Publication},
  note         = {Official report published by the Statistics and Informatics Division; accessed 2026-03-28}
}

@article{chang2024goldfish,
  title   = {Goldfish: Monolingual Language Models for 350 Languages},
  author  = {Chang, Tyler A. and Arnett, Catherine and Tu, Zhuowen and Bergen, Benjamin K.},
  journal = {arXiv preprint arXiv:2408.10441},
  year    = {2024},
  url     = {https://arxiv.org/abs/2408.10441}
}

@misc{dailystar2025internet,
  title        = {Internet Shows Stark Rural-Urban Divide},
  author       = {{The Daily Star}},
  year         = {2025},
  month        = jan,
  day          = {1},
  url          = {https://www.thedailystar.net/business/news/internet-shows-stark-rural-urban-divide-3789861},
  note         = {Reporting on BBS ICT Survey; accessed 2026-03-28}
}

@article{dettmers2023qlora,
  title   = {{QLoRA}: Efficient Finetuning of Quantized {LLMs}},
  author  = {Dettmers, Tim and Pagnoni, Artidoro and Holtzman, Ari and Zettlemoyer, Luke},
  journal = {Advances in Neural Information Processing Systems},
  volume  = {36},
  pages   = {10088--10115},
  year    = {2023},
  doi     = {10.48550/arXiv.2305.14314},
  url     = {https://arxiv.org/abs/2305.14314}
}

@article{dodoo2025teacher,
  title   = {The Teacher's Dilemma: Balancing Trade-Offs in Programming Education for Emergent Bilingual Students},
  author  = {Dodoo, Emma R. and Nelson-Fromm, Tamara and Guzdial, Mark},
  journal = {arXiv preprint arXiv:2506.14147},
  year    = {2025},
  url     = {https://arxiv.org/abs/2506.14147}
}

@misc{ethnologue2025,
  title        = {What Are the Top 200 Most Spoken Languages?},
  author       = {{Ethnologue}},
  year         = {2025},
  howpublished = {Ethnologue Insights},
  url          = {https://www.ethnologue.com/insights/ethnologue200//},
  note         = {Accessed 2026-03-28}
}

@inproceedings{hasan2021xlsum,
  title     = {{XL-Sum}: Large-Scale Multilingual Abstractive Summarization for 44 Languages},
  author    = {Hasan, Tahmid and Bhattacharjee, Abhik and Islam, Md. Saiful and Mubasshir, Kazi and Li, Yuan-Fang and Kang, Yong-Bin and Rahman, M. Sohel and Shahriyar, Rifat},
  booktitle = {Findings of the Association for Computational Linguistics: ACL-IJCNLP 2021},
  pages     = {4693--4703},
  year      = {2021},
  publisher = {Association for Computational Linguistics},
  url       = {https://aclanthology.org/2021.findings-acl.413/}
}

@inproceedings{hu2022lora,
  title     = {{LoRA}: Low-Rank Adaptation of Large Language Models},
  author    = {Hu, Edward J. and Shen, Yelong and Wallis, Phillip and Allen-Zhu, Zeyuan and Li, Yuanzhi and Wang, Shean and Wang, Lu and Chen, Weizhu},
  booktitle = {International Conference on Learning Representations ({ICLR})},
  year      = {2022},
  url       = {https://arxiv.org/abs/2106.09685}
}

@misc{icls2026,
  title        = {10 Most Spoken Languages in the World in 2026},
  author       = {{International Communication and Leadership School}},
  year         = {2026},
  url          = {https://www.icls.edu/blog/most-spoken-languages-in-the-world},
  note         = {Accessed 2026-03-28}
}

@inproceedings{joshi2020state,
  title     = {The State and Fate of Linguistic Diversity and Inclusion in the {NLP} World},
  author    = {Joshi, Pratik and Santy, Sebastin and Budhiraja, Amar and Bali, Kalika and Choudhury, Monojit},
  booktitle = {Proceedings of the 58th Annual Meeting of the Association for Computational Linguistics},
  pages     = {6282--6293},
  year      = {2020},
  publisher = {Association for Computational Linguistics},
  url       = {https://aclanthology.org/2020.acl-main.560/}
}

@inproceedings{kabir2024benllm,
  title     = {{BenLLM-Eval}: A Comprehensive Evaluation into the Potentials and Pitfalls of Large Language Models on {Bengali} {NLP}},
  author    = {Kabir, Mohsinul and Islam, Mohammed Saidul and Laskar, Md Tahmid Rahman and Nayeem, Mir Tafseer and Bari, M. Saiful and Hoque, Enamul},
  booktitle = {Proceedings of the 2024 Joint International Conference on Computational Linguistics, Language Resources and Evaluation ({LREC-COLING})},
  pages     = {2238--2252},
  year      = {2024},
  url       = {https://aclanthology.org/2024.lrec-main.201/}
}

@inproceedings{khan2024sangraha,
  title     = {{IndicLLMSuite}: A Blueprint for Creating Pre-training and Fine-Tuning Datasets for Indian Languages},
  author    = {Khan, Mohammed Safiur Rahman and Madhani, Yash and Nilesh, Vinit and others},
  booktitle = {Proceedings of the 62nd Annual Meeting of the Association for Computational Linguistics (Volume 1: Long Papers)},
  pages     = {9743--9752},
  year      = {2024},
  publisher = {Association for Computational Linguistics},
  url       = {https://aclanthology.org/2024.acl-long.843/},
  note      = {Describes the Sangraha data pipeline}
}

@misc{langlais2025commoncorpus,
  title        = {Common Corpus: The Largest Collection of Ethical Data for {LLM} Pre-training},
  author       = {Langlais, Pierre-Jean and others},
  year         = {2025},
  howpublished = {Hugging Face Dataset},
  url          = {https://huggingface.co/datasets/PleIAs/common_corpus},
  note         = {Dataset documentation available at arXiv:2506.01732; accessed 2026-03-28}
}

@misc{pmindia2024classical,
  title        = {Cabinet Approves Conferring Status of Classical Language to {Marathi}, {Pali}, {Prakrit}, {Assamese} and {Bengali} Languages},
  author       = {{Prime Minister's Office, Government of India}},
  year         = {2024},
  month        = oct,
  day          = {3},
  url          = {https://www.pmindia.gov.in/en/news_updates/cabinet-approves-conferring-status-of-classical-language-to-marathi-pali-prakrit-assamese-and-bengali-languages/},
  note         = {Accessed 2026-03-28}
}

@misc{pimienta2024,
  title        = {Reliably Exploring the Presence of Languages on the Internet},
  author       = {Pimienta, Daniel},
  year         = {2024},
  howpublished = {Research Outreach},
  doi          = {10.32907/RO-139-5856605838},
  url          = {https://doi.org/10.32907/RO-139-5856605838},
  note         = {Accessed 2026-03-28}
}

@inproceedings{raihan2025blp,
  title     = {Overview of {BLP}-2025 Task 2: Code Generation in {Bangla}},
  author    = {Raihan, Nafees and Jawad, Mohammad Abdullah and Rahman, Md. Mosiur and others},
  booktitle = {Proceedings of the Second Workshop on {Bangla} Language Processing ({BLP}-2025)},
  pages     = {373--387},
  year      = {2025},
  publisher = {Association for Computational Linguistics},
  url       = {https://aclanthology.org/2025.banglalp-1.31/}
}

@inproceedings{rashed2025bridging,
  title   = {Bridging the Last Mile: Unpacking the Rural Digital Divide in {Bangladesh}},
  author  = {Rashed, Rayhan and Bhuyan, Muhammad Masroor Ali and others},
  booktitle = {Proceedings of the 2025 {ACM SIGCAS/SIGCHI} Conference on Computing and Sustainable Societies ({COMPASS})},
  pages   = {150--166},
  year    = {2025},
  doi     = {10.1145/3715335.3735463},
  url     = {https://doi.org/10.1145/3715335.3735463}
}

@article{roussel2017cognitiveload,
  title   = {Learning Subject Content through a Foreign Language Should Not Ignore Human Cognitive Architecture: A Cognitive Load Theory Approach},
  author  = {Roussel, St{\'e}phanie and Joulia, Danielle and Tricot, Andr{\'e} and Sweller, John},
  journal = {Learning and Instruction},
  volume  = {52},
  pages   = {69--79},
  year    = {2017},
  doi     = {10.1016/j.learninstruc.2017.04.007},
  url     = {https://doi.org/10.1016/j.learninstruc.2017.04.007}
}

@inproceedings{shahriar2024improving,
  title     = {Improving {Bengali} and {Hindi} Large Language Models},
  author    = {Shahriar, Arif and Barbosa, Denilson},
  booktitle = {Proceedings of the 2024 Joint International Conference on Computational Linguistics, Language Resources and Evaluation ({LREC-COLING})},
  pages     = {8776--8787},
  year      = {2024},
  url       = {https://aclanthology.org/2024.lrec-main.764/}
}

@inproceedings{soosairaj2018native,
  title     = {Does Native Language Play a Role in Learning a Programming Language?},
  author    = {Soosai Raj, Arumugam Godfrey and Ketsuriyonk, Kanjanee and Patel, Jignesh M. and Halverson, Richard},
  booktitle = {Proceedings of the 49th {ACM} Technical Symposium on Computer Science Education ({SIGCSE} '18)},
  pages     = {11--16},
  year      = {2018},
  doi       = {10.1145/3159450.3159531},
  url       = {https://doi.org/10.1145/3159450.3159531}
}

@book{sweller2011clt,
  title     = {Cognitive Load Theory},
  author    = {Sweller, John and Ayres, Paul and Kalyuga, Slava},
  publisher = {Springer},
  address   = {New York, NY},
  year      = {2011},
  doi       = {10.1007/978-1-4419-8126-4},
  url       = {https://doi.org/10.1007/978-1-4419-8126-4}
}

@misc{tbs2025internet,
  title        = {52\% Households in {Bangladesh} Have Internet Connections: {BBS}},
  author       = {{The Business Standard}},
  year         = {2025},
  month        = apr,
  day          = {12},
  url          = {https://www.tbsnews.net/tech/52-households-bangladesh-have-internet-connections-bbs-1114086},
  note         = {Accessed 2026-03-28}
}

@misc{un_imld,
  title        = {International Mother Language Day},
  author       = {{United Nations}},
  year         = {2026},
  url          = {https://www.un.org/en/observances/mother-language-day},
  note         = {Accessed 2026-03-28}
}

@misc{unesco_imld,
  title        = {International Mother Language Day},
  author       = {{UNESCO}},
  year         = {2026},
  url          = {https://www.unesco.org/en/days/mother-language},
  note         = {Accessed 2026-03-28}
}

@misc{w3techs2026,
  title        = {Usage Statistics and Market Share of Content Languages for Websites, March 2026},
  author       = {{W3Techs}},
  year         = {2026},
  url          = {https://w3techs.com/technologies/overview/content_language},
  note         = {English listed at 49.5\% on 2026-03-28; accessed 2026-03-28}
}

\end{document}